\documentclass[10pt,onecolumn]{wlscirep}
\usepackage{xcolor}
\usepackage{titlesec}
\usepackage{parskip}
\usepackage{graphicx}
\usepackage{psfrag}
\usepackage{subfigure}
\usepackage{rotating}
\usepackage{tabularx}
\usepackage{siunitx}
\usepackage{amssymb}
\usepackage{float}
\usepackage{stfloats}
\usepackage{hhline}
\usepackage{mathtools}
\titlespacing{\section}{0pt}{\parskip}{5pt}
\titlespacing{\subsection}{0pt}{\parskip}{5pt}
\title{Regularization of Soft Actor-Critic Algorithms with Automatic Temperature Adjustment}

\author[1*]{Ben You}

\affil[1*]{you.ben@polymtl.ca}


\begin{abstract}
This work presents a comprehensive analysis to regularize the Soft Actor-Critic (SAC) algorithm with automatic temperature adjustment \cite{haarnoja2018soft, haarnoja2018learning}. The the policy evaluation, the policy improvement and the temperature adjustment are reformulated, addressing certain modification and enhancing the clarity of the original theory in a more explicit manner.
\end{abstract}

\begin{document}
\flushbottom
\maketitle

\section{Introduction}

The Soft Actor-Critic (SAC) algorithm with automatic temperature adjustment \cite{haarnoja2018soft,haarnoja2018learning}is an extension of the original SAC algorithm \cite{haarnoja2018soft1} that incorporates a mechanism to automatically adjust the temperature parameter. The SAC algorithm has demonstrated strong performance on a wide range of reinforcement learning tasks, including robotic control, continuous locomotion, and manipulation tasks. It achieves state-of-the-art performance and is known for its stability, sample efficiency, and ability to handle high-dimensional continuous action spaces. However, as with any algorithm, the performance of SAC can be influenced by the choice of hyperparameters, network architecture, and the complexity of the task at hand.

Since the introduction of the automatic temperature version of the SAC algorithm came after the fixed temperature version, there may be some ambiguity in the development of the theory, particularly in the derivation of the recursive definition of the soft-Q function. In this work, a thorough deduction of the Bellmann equation for soft-Q function, policy improvement and automatic temperature adjustment is presented, so as to clarify the ambiguities and correct the defects found in the original articles\cite{haarnoja2018soft,haarnoja2018learning}.

\section{Recursive definition of soft-Q function}
In the original paper \cite{haarnoja2018soft}, the optimization problem of maximizing the reward under the constraint of a lower-bound entropy is formulated as follows (without loss of generality, the discount factor $\gamma$ is set to 1):
\begin{equation}\label{eq:original opt problem}
    \mathop{\max}_{\pi_{0 \colon T}}\mathop{\mathbb{E}}\left[\sum^{T}_{t=0} r(s_{t},a_{t})\right] \ s.t. \mathop{\mathbb{E}}_{(s_{t},a_{t})\sim \rho_{\pi_{t}}}\left[-\log(\pi_{t}(a_{t}\mid s_{t}))\right]  \geqslant H_{0} \ (\forall t=0,\cdots,T)
\end{equation}
where $\rho_{\pi_{t}}$ denote the state-action marginal of the trajectory distribution induced by a policy $\pi_{t}$.

Since the policy at time $t$ can only affect the future objective value, we can rewrite it as a dynamic programming form:
\begin{equation}\label{eq:DP}
\mathop{\max}_{\pi_{0}}\left({\mathbb{E}}_{\rho_{\pi_{0}}}[r(s_{0},a_{0})]+\cdots+\mathop{\max}_{\pi_{T-1}}\left({\mathbb{E}}_{\rho_{\pi_{T-1}}}[r(s_{T-1},a_{T-1})]+\mathop{\max}_{\pi_{T}}\left({\mathbb{E}}_{\rho_{\pi_{T}}}[r(s_{T},a_{T})]\right)\right)\cdots\right)
\end{equation}
with constraints $\mathop{\mathbb{E}}_{ \rho_{\pi_{t}}}\left[-\log(\pi_{t}(a_{t}\mid s_{t}))\right]  \geqslant H_{0} \ (\forall t=0,\cdots,T)$.

For the sake of concise expression, we denote some variables as
\begin{enumerate}
    \vspace{5pt}
    \item $r_{t}=r(s_{t},a_{t})$
    
    \item $h(\pi_{t})=\mathop{\mathbb{E}}_{ \rho_{\pi_{t}}}\left[-\log(\pi_{t}(a_{t}\mid s_{t}))\right]-H_{0}$
    \vspace{5pt}
\end{enumerate}
Consequently, \eqref{eq:DP} can be expressed as
\begin{equation}
\mathop{\max}_{\substack{\pi_{0} \\ h(\pi_{0})  \geqslant 0}}\left({\mathbb{E}}_{\rho_{\pi_{0}}}[r_{0}]+\cdots+\mathop{\max}_{\substack{\pi_{T-1} \\ h(\pi_{T-1})  \geqslant 0}}\left({\mathbb{E}}_{\rho_{\pi_{T-1}}}[r_{T-1}]+\mathop{\max}_{\substack{\pi_{T} \\ h(\pi_{T})  \geqslant 0}}\left({\mathbb{E}}_{\rho_{\pi_{T}}}[r_{T}]\right)\right)\cdots\right)
\end{equation}

\subsection{Step T}
In the step $T$, the corresponding optimization problem is given by
\begin{equation}\label{eq:step T}
\mathrm{p}^{*}_{T}=\mathop{\max}_{\substack{\pi_{T} \\ h(\pi_{T})  \geqslant 0}}\left({\mathbb{E}}_{\rho_{\pi_{T}}}[r_{T}]\right)
\end{equation}
In turn, we can derive the corresponding Lagrangian as
\begin{equation}
    L_{T}(\pi_{T}, \alpha_{T})={\mathbb{E}}_{\rho_{\pi_{T}}}[r_{T}]+\alpha_{T} h(\pi_{T})
\end{equation}
\\
Because the prime problem \eqref{eq:step T} is concave and Slater's condition holds, we have strong duality:
\begin{equation}
\mathrm{p}^{*}_{T}=\mathrm{d}^{*}_{T}
\end{equation}
where $\mathrm{d}^{*}_{T}$ is given by
\begin{equation}\label{eq:dT}
\mathrm{d}^{*}_{T}=\mathop{\min}_{\alpha_{T}\geqslant 0}\mathop{\max}_{\pi_{T}}L_{T}(\pi_{T}, \alpha_{T})=\mathop{\min}_{\alpha_{T}\geqslant 0}\mathop{\max}_{\pi_{T}}{{\mathbb{E}}_{\rho_{\pi_{T}}}[r_{T}]}+\alpha_{T} h(\pi_{T})
\end{equation}
Or equivalently, we have
\begin{equation}\label{eq:dT optimal}
\mathrm{d}^{*}_{T}=\textcolor{teal}{{\mathbb{E}}_{\rho_{\pi^{*}_{T}}}[r_{T}]}+\alpha^{*}_{T} h(\pi^{*}_{T})
\end{equation}
\\
The corresponding optimal variables ${\pi}^{*}_{T}$ and ${\alpha}^{*}_{T}$ are respectively given by
\begin{equation}
    {\pi}^{*}_{T}=\mathop{\mathrm{argmax}}_{\pi_{T}}{\mathbb{E}}_{\rho_{\pi_{T}}}[r_{T}]+\alpha_{T} h(\pi_{T})
\end{equation}
and
\begin{equation}
    {\alpha}^{*}_{T}=\mathop{\mathrm{argmin}}_{\alpha_{T}\geqslant 0}\alpha_{T} h(\pi^{*}_{T})
\end{equation}

\subsection{Step T-1}
\begin{equation}\label{eq:step T-1}
\mathrm{p}^{*}_{T-1}=\mathop{\max}_{\substack{\pi_{T-1} \\ h(\pi_{T-1})  \geqslant 0}}\left({\mathbb{E}}_{\rho_{\pi_{T-1}}}[r_{T-1}]+\mathop{\max}_{\substack{\pi_{T} \\ h(\pi_{T})  \geqslant 0}}\left({\mathbb{E}}_{\rho_{\pi_{T}}}[r_{T}]\right)\right)
\end{equation}
\\
Plugging \eqref{eq:dT optimal} to \eqref{eq:step T-1}, we have
\begin{equation}\label{eq:orginal dTm1}
\mathrm{p}^{*}_{T-1}=\mathop{\max}_{\substack{\pi_{T-1} \\ h(\pi_{T-1})  \geqslant 0}}\left({\mathbb{E}}_{\rho_{\pi_{T-1}}}[r_{T-1}]+{{\mathbb{E}}_{\rho_{\pi^{*}_{T}}}[r_{T}]}+\alpha^{*}_{T} h(\pi^{*}_{T})\right)
\end{equation}
Again, applying strong duality on \eqref{eq:orginal dTm1}, the optimal duality $\mathrm{d}^{*}_{T-1}$ can be obtained as
\begin{equation}
    \mathrm{d}^{*}_{T-1}=\mathop{\min}_{\alpha_{T-1}\geqslant 0}\mathop{\max}_{\pi_{T-1}}{{{\mathbb{E}}_{\rho_{\pi_{T-1}}}[r_{T-1}]}+{{\mathbb{E}}_{\rho_{\pi^{*}_{T}}}[r_{T}]}+\alpha^{*}_{T} h(\pi^{*}_{T})}+\alpha_{T-1} h(\pi_{T-1})
\end{equation}
Or equivalently, 
\begin{equation}\label{eq:dTm1 optimal}
\mathrm{d}^{*}_{T-1}=\textcolor{blue}{{\mathbb{E}}_{\rho_{\pi^{*}_{T-1}}}[r_{T-1}]}+\textcolor{teal}{{{\mathbb{E}}_{\rho_{\pi^{*}_{T}}}[r_{T}]}}+\textcolor{blue}{\alpha^{*}_{T} h(\pi^{*}_{T})}+\alpha^{*}_{T-1} h(\pi^{*}_{T-1})
\end{equation}
\\
Similar to the case of step $T$, the optimal variables ${\pi}^{*}_{T-1}$ and ${\alpha}^{*}_{T-1}$ are respectively given by
\begin{equation}
    {\pi}^{*}_{T-1}=\mathop{\mathrm{argmax}}_{\pi_{T-1}}{{{\mathbb{E}}_{\rho_{\pi_{T-1}}}[r_{T-1}]}+{{\mathbb{E}}_{\rho_{\pi^{*}_{T}}}[r_{T}]}+\alpha^{*}_{T} h(\pi^{*}_{T})}+\alpha_{T-1} h(\pi_{T-1})
\end{equation}
and
\begin{equation}
    {\alpha}^{*}_{T-1}=\mathop{\mathrm{argmin}}_{\alpha_{T-1}\geqslant 0}\alpha_{T-1} h(\pi^{*}_{T-1})
\end{equation}    

\subsection{Step T-2}
Based on Section 1 and Section 2, it is straightforward to derive the strong duality $\mathrm{d}^{*}_{T-2}$ for the case of step $T-2$ as shown below:
\begin{equation}\label{eq:dTm2 optimal}
\mathrm{d}^{*}_{T-2}=\textcolor{cyan}{{\mathbb{E}}_{\rho_{\pi^{*}_{T-2}}}[r_{T-2}]}+\textcolor{blue}{{\mathbb{E}}_{\rho_{\pi^{*}_{T-1}}}[r_{T-1}]}+\textcolor{teal}{{{\mathbb{E}}_{\rho_{\pi^{*}_{T}}}[r_{T}]}}+\textcolor{blue}{\alpha^{*}_{T} h(\pi^{*}_{T})}+\textcolor{cyan}{\alpha^{*}_{T-1} h(\pi^{*}_{T-1})}+\alpha^{*}_{T-2} h(\pi^{*}_{T-2})
\end{equation}
And the optimal variables $\alpha^{*}_{T-2}$ and $\pi^{*}_{T-2}$ are respectively given by
\begin{equation}
    {\pi}^{*}_{T-2}=\mathop{\mathrm{argmax}}_{\pi_{T-2}}{{\mathbb{E}}_{\rho_{\pi_{T-2}}}[r_{T-2}]}+{{\mathbb{E}}_{\rho_{\pi^{*}_{T-1}}}[r_{T-1}]}+{{{\mathbb{E}}_{\rho_{\pi^{*}_{T}}}[r_{T}]}}+{\alpha^{*}_{T} h(\pi^{*}_{T})}+{\alpha^{*}_{T-1} h(\pi^{*}_{T-1})}+\alpha_{T-2} h(\pi_{T-2})
\end{equation}
and
\begin{equation}
    {\alpha}^{*}_{T-2}=\mathop{\mathrm{argmin}}_{\alpha_{T-2}\geqslant 0}\alpha_{T-2} h(\pi^{*}_{T-2})
\end{equation}

\subsection{Step t}
Based on \eqref{eq:dT optimal}, \eqref{eq:dTm1 optimal} and \eqref{eq:dTm2 optimal},  we can establish the following set of equations as an iterative process:
\begin{equation}
    \begin{array}{lc}
        \vspace{5pt}
         \textcolor{teal}{\Bar{Q}_{T}}=\textcolor{teal}{{\mathbb{E}}_{\rho_{\pi_{T}}}[r_{T}]}\\
         \vspace{5pt}
         \textcolor{blue}{\Bar{Q}_{T-1}}=\textcolor{blue}{{\mathbb{E}}_{\rho_{\pi_{T-1}}}[r_{T-1}]}+\textcolor{teal}{\Bar{Q}_{T}}+\textcolor{blue}{\alpha_{T} h(\pi_{T})}\\
         \vspace{5pt}
         \textcolor{cyan}{\Bar{Q}_{T-2}}=\textcolor{cyan}{{\mathbb{E}}_{\rho_{\pi_{T-2}}}[r_{T-2}]}+\textcolor{blue}{\Bar{Q}_{T-1}}+\textcolor{cyan}{\alpha_{T-1} h(\pi_{T-1})}\\
         \vspace{5pt}
         \cdots \\
         \vspace{5pt}
         {\Bar{Q}_{t}}={{\mathbb{E}}_{\rho_{\pi_{t}}}[r_{t}]}+\left[{\Bar{Q}_{t+1}}+{\alpha_{t+1} h(\pi_{t+1})}\right]\\
         \vspace{5pt}
         \cdots \\
    \end{array}
\end{equation}
For the sake of comparison, we expand the expression of the above recursive equation as follows:
\begin{equation}\label{eq: qt recursive}
 \Bar{Q}_{t}=\mathop{\mathbb{E}}_{(s_{t},a_{t})\sim \rho_{\pi_{t}}}[r(s_{t},a_{t})]+ \Bar{Q}_{t+1}+\mathop{\mathbb{E}}_{(s_{t+1},a_{t+1})\sim \rho_{\pi_{t+1}}}[-\alpha_{t+1}\log(\pi_{t+1}(a_{t+1}\mid s_{t+1}))-\alpha_{t+1}H_{0}]
\end{equation}
\section{Revision of Bellmann equation for soft-Q function}
Inspired by the expression of soft Q-value Bellmann equation\cite{haarnoja2018soft}:
\begin{equation}\label{eq:soft q-value Bellemann}
    Q(s_{t},a_{t})=r(s_{t},a_{t})+\mathop{\mathbb{E}}_{s_{t+1}\sim p,\ a_{t+1}\sim \pi_{t+1}}[Q(s_{t+1},a_{t+1})-\alpha\log(\pi_{t+1}(a_{t+1}\mid s_{t+1}))]
\end{equation}
We take $\Bar{Q}_{t}$ as an expectation:
\begin{equation}\label{eq:Qt}
    \Bar{Q}_{t}=\mathop{\mathbb{E}}_{(s_{t},a_{t})\sim \rho_{\pi_{t}}}[Q(s_{t},a_{t})]
\end{equation}
Inserting \eqref{eq:Qt} to \eqref{eq: qt recursive}, after some deduction, we have
\begin{equation} \label{eq:modified q-value Bellmann}
    Q(s_{t},a_{t})=r(s_{t},a_{t})+\mathop{\mathbb{E}}_{s_{t+1}\sim p,\ a_{t+1}\sim \pi_{t+1}}[Q(s_{t+1},a_{t+1})-\alpha_{t+1}\log\pi_{t+1}(a_{t+1}\mid s_{t+1})\textcolor{red}{-\alpha_{t+1}H_{0}}]
\end{equation}
where $p$ represents the transition probability $P(s_{t+1}\mid s_{t}, a_{t})$. Note that \eqref{eq:modified q-value Bellmann} is the special solution of '$E[X]=0$ when $X=0$'. Besides the item related to $H_{0}$, \eqref{eq: qt recursive} is identical to \eqref{eq:soft q-value Bellemann}, which can be referred to as the soft Q-value Bellmann equation with lower-bounded entropy.

To facilitate comparison, we present the recursive definition of the soft Q-function (Eqn.(15)\cite{haarnoja2018soft}):
\begin{equation} \label{eq:soft recursive Q-value Bellemann}
    Q^{*}_{t}(s_{t},a_{t};\pi^{*}_{t+1:T},\alpha^{*}_{t+1:T})=\mathop{\mathbb{E}}[r(s_{t},a_{t})]+\mathbb{E}_{ \rho_{\pi}}\left[Q^{*}_{t+1}(s_{t+1},a_{t+1})-\alpha^{*}_{t+1}\log(\pi^{*}_{t+1}(a_{t+1}\mid s_{t+1}))\right]
\end{equation}
By meticulously examining \eqref{eq:soft recursive Q-value Bellemann} alongside \eqref{eq: qt recursive} and \eqref{eq:modified q-value Bellmann}, it becomes evident that the expression is incorrect. In fact, it can be considered a conflation of the latter two equations, leading to confusion.

\subsection{Discussion of the absence of target entropy}
The absence of $H_{0}$ in \eqref{eq:soft q-value Bellemann} may pose a problem during the training process. Throughout the following discussion, we will consistently use \eqref{eq: qt recursive} as our benchmark for comparison. Note that given $|A|<\infty$, the entropy of uniform distribution is the upper bound of the differential entropy, i.e., $H(\pi(\textbf{a}\mid \textbf{s}))\leq log{|A|}$, where $|A|=\prod_{n=1}^{dim(A)} \max(a_{n})-\min(a_{n})$.



\begin{figure}[!hbtp]    \centering\includegraphics[width=14cm]{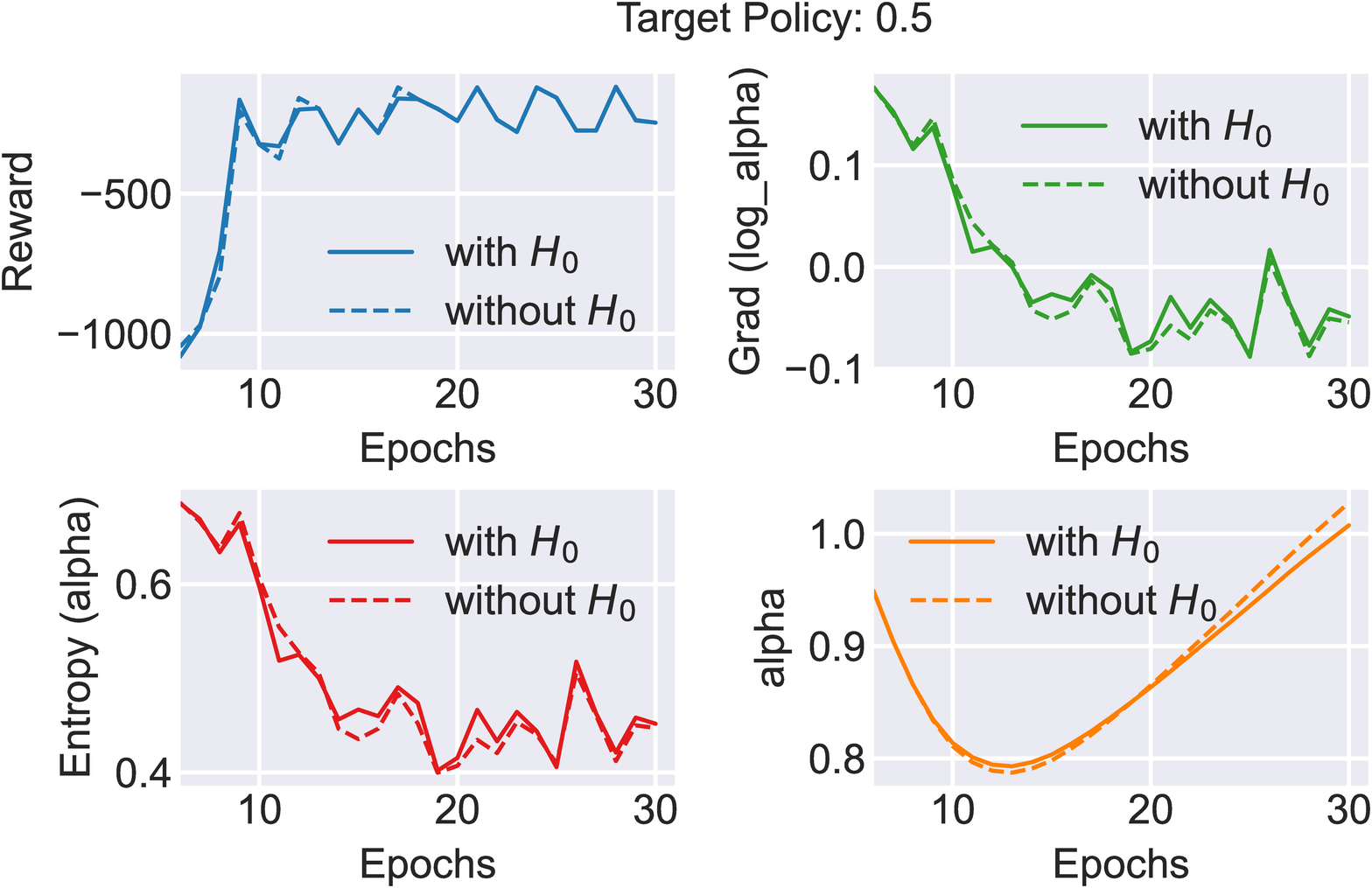}
    \caption{Case of Pendulum-v1. The target entropy $H_{0}$ and the initial $\alpha_{0}$ are set to 0.5 and 1, respectively. Both results are obtained using a fixed random seed. It's important to note that the logarithmic temperature is updated using the SGD algorithm with zero weight decay, ensuring that the rise or fall of $\alpha$ is solely determined by the sign of $h(\pi^{*}_{t})$.}
    \label{fig: target entropy: 0.5}
\end{figure}

In the case of $0<H_{0}\leq log{|A|}$, using \eqref{eq:soft q-value Bellemann} to update the Q function results in overestimation, leading to a sharper Boltzmann distribution. Consequently, the policy network generates a corresponding Gaussian distribution with lower entropy. When the original value of $h(\pi^{*}_{t})$ is negative, reducing the entropy leads to a larger magnitude of $|h(\pi^{*}_{t})|$, which increases the temperature. When the original value of $h(\pi^{*}_{t})$ is positive, the absence of $H_{0}$ hinders the reduction of entropy. Moreover, there is a possibility that $h(\pi^{*}_{t})$ becomes negative, causing an increase in $\alpha^{*}_{t}$. In sum, the absence of $H_{0}$ pushes the SAC algorithm towards over-exploration, as illustrated in Fig.\ref{fig: target entropy: 0.5}.

Conversely, when $H_{0}<0$, utilizing \eqref{eq:soft q-value Bellemann} to update the Q function leads to underestimation, resulting in a flatter Boltzmann distribution. As a result, the policy network generates a Gaussian distribution with higher entropy. Similar to the previous scenario, we can deduce that the absence of $H_{0}$ leads to the issue of under-exploration.

As the empirical value of $H_{0}$ is often set to $-dim(A)$ \cite{haarnoja2018soft}, it suggests that the absence of $H_{0}$ in the Bellman backup equation for the Q function might fall into the aforementioned under-exploration. 



\section{Policy improvement and automatic temperature adjustment}
The optimal $\pi^{*}_{t}$ can be expressed as
\begin{equation}
\pi^{*}_{t}=\mathop{\mathrm{argmax}}_{\pi_{t}} \Bar{Q}_{t}+\alpha_{t} h(\pi_{t})
=\mathop{\mathrm{argmax}}_{\pi_{t}}\mathop{\mathbb{E}}_{(s_{t},a_{t})\sim \rho_{\pi_{t}}}[Q(s_{t},a_{t})-\alpha_{t}(\log\pi_{t}(a_{t}\mid s_{t})+H_{0})]
\end{equation}
Or equivalently,
\begin{equation}\label{eq:policy improvement}
    \pi^{*}_{t}=\mathop{\mathrm{argmin}}_{\pi_{t}}\mathop{\mathbb{E}}_{\textcolor{red}{s_{t}\sim \rho_{\pi_{t}}},\ a_{t}\sim \pi_{t}}[\alpha_{t}\log\pi_{t}(a_{t}\mid s_{t})-Q(s_{t},a_{t})]
\end{equation}
Comparing to Eqn.(4) \cite{haarnoja2018soft1}, \eqref{eq:policy improvement} includes an additional expectation with respect to $s_{t}$. It is worth noting that this expectation operator is already incorporated in Eqn.(7) \cite{haarnoja2018soft1}, which defines the loss function of the policy network (replay buffer is implemented). However, it is important to clarify that the author's claim stating 'While in principle we could choose any projection, it will
turn out to be convenient to use the information projection defined in terms of the Kullback-Leibler
divergence.' is incorrect. This is because the policy improvement, which is explicitly dependent on \eqref{eq:policy improvement}, arises from the optimization problem stated in \eqref{eq:original opt problem}. Furthermore, when introducing the Boltzmann distribution \cite{haarnoja2018soft1, haarnoja2018soft, haarnoja2018learning}, it is crucial to ensure that the 'energy' satisfies a non-negative presumption. Specifically, this implies that the Q-value function should be non-positive. However, it is worth noting that \eqref{eq:policy improvement} does not possess such a constraint.

Correspondingly, the optimal $\alpha^{*}_{t}$ can be expressed as
\begin{equation}
    {\alpha}^{*}_{t}=\mathop{\mathrm{argmin}}_{\alpha_{t}\geqslant 0}\alpha_{t} h(\pi^{*}_{t})=\mathop{\mathrm{argmin}}_{\alpha_{t}\geqslant 0}\alpha_{t}\left\{\mathop{\mathbb{E}}_{\textcolor{red}{s_{t}\sim \rho_{\pi^{*}_{t}}},\ a_{t}\sim \pi^{*}_{t}}-[\log\pi^{*}_{t}(a_{t}\mid s_{t})+H_{0}]\right\}
\end{equation}
Similarly, when comparing to Eqn.(17)\cite{haarnoja2018soft}, the only difference is the inclusion of an expectation with respect to $s_{t}$. Note that in the corresponding Python code for the loss function of the temperature in the SAC algorithm, the mean value of samples from the replay buffer is utilized to approximate the expectations. Therefore, the contribution of the expectation with respect to $s_{t}$ is already accounted for in the implementation.

\section{Conclusion}
In conclusion, we have successfully demonstrated the incorrectness of the recursive definition of the soft-Q function presented in the article "Soft Actor-Critic Algorithms and Applications" \cite{haarnoja2018soft}. There exists a missing item $-\alpha H_{0}$ in the expression of Bellmann backup operator which might induce over-/under-exploration in the policy evaluation process. Moreover, the policy improvement is determined by the optimization problem \eqref{eq:original opt problem} rather than arbitrary information projection. Last but not least, the policy improvement and automatic temperature adjustment must incorporate the expectation with respect to the state. 

\bibliography{Reference}

\begin{thebibliography}{1}
\urlstyle{rm}
\expandafter\ifx\csname url\endcsname\relax
  \def\url#1{\texttt{#1}}\fi
\expandafter\ifx\csname urlprefix\endcsname\relax\def\urlprefix{URL }\fi
\expandafter\ifx\csname doiprefix\endcsname\relax\def\doiprefix{DOI: }\fi
\providecommand{\bibinfo}[2]{#2}
\providecommand{\eprint}[2][]{\url{#2}}

\bibitem{haarnoja2018soft}
\bibinfo{author}{Haarnoja, T.} \emph{et~al.}
\newblock \bibinfo{journal}{\bibinfo{title}{Soft actor-critic algorithms and
  applications}}.
\newblock {\emph{\JournalTitle{arXiv preprint arXiv:1812.05905}}}
  (\bibinfo{year}{2018}).

\bibitem{haarnoja2018learning}
\bibinfo{author}{Haarnoja, T.} \emph{et~al.}
\newblock \bibinfo{journal}{\bibinfo{title}{Learning to walk via deep
  reinforcement learning}}.
\newblock {\emph{\JournalTitle{arXiv preprint arXiv:1812.11103}}}
  (\bibinfo{year}{2018}).

\bibitem{haarnoja2018soft1}
\bibinfo{author}{Haarnoja, T.}, \bibinfo{author}{Zhou, A.},
  \bibinfo{author}{Abbeel, P.} \& \bibinfo{author}{Levine, S.}
\newblock \bibinfo{title}{Soft actor-critic: Off-policy maximum entropy deep
  reinforcement learning with a stochastic actor}.
\newblock In \emph{\bibinfo{booktitle}{International conference on machine
  learning}}, \bibinfo{pages}{1861--1870} (\bibinfo{organization}{PMLR},
  \bibinfo{year}{2018}).

\end{thebibliography}

\end{document}